\documentclass[conference]{IEEEtran}
\ifCLASSINFOpdf \else \fi
\usepackage{cite}
\ifCLASSOPTIONcompsoc
 \usepackage[caption=false,font=normalsize,labelfont=sf,textfont=sf]{subfig}
\else
 \usepackage[caption=false,font=footnotesize]{subfig}
\fi
\usepackage{amsmath}%,amssymb,amsfonts}
\usepackage{graphicx}
\usepackage{xcolor}
\def\BibTeX{{\rm B\kern-.05em{\sc i\kern-.025em b}\kern-.08em
    T\kern-.1667em\lower.7ex\hbox{E}\kern-.125emX}}
\usepackage{hyperref}
\usepackage{cite}
\usepackage{amsmath}
\usepackage{graphicx}
\usepackage{float}
\ifCLASSOPTIONcompsoc
    \usepackage[caption=false, font=normalsize, labelfont=sf, textfont=sf]{subfig}
\else
\usepackage[caption=false, font=footnotesize]{subfig}
\fi

\usepackage{svg} % .svg dear jonno
\usepackage{multirow} % for multiple row cells in a table
\usepackage{verbatim} % 

\usepackage{url}
\usepackage{hyperref}
\hypersetup{
    colorlinks = true,
    linkcolor  = red,
    filecolor = magenta,      
    urlcolor = cyan,
}

\makeatletter
 \let\old@ps@headings\ps@headings
 \let\old@ps@IEEEtitlepagestyle\ps@IEEEtitlepagestyle
 \def\confheader#1{%
 \def\ps@IEEEtitlepagestyle{%
 \old@ps@IEEEtitlepagestyle%
 \def\@oddhead{\strut\hfill#1\hfill\strut}%
 \def\@evenhead{\strut\hfill#1\hfill\strut}%
 }%
 \ps@headings%
 }
 \makeatother

\confheader{%
 2nd ICAICT, 28-29 November 2020, Dhaka, Bangladesh.
 }

 \usepackage[pscoord]{eso-pic}
\newcommand{\placetextbox}[3]{
 \setbox0=\hbox{#3}
 \AddToShipoutPictureFG*{ \put(\LenToUnit{#1\paperwidth},\LenToUnit{#2\paperheight}){\vtop{{\null}\makebox[0pt][c]{#3}}}
 }
 }
 \placetextbox{.23}{0.055}{\small{978-0-7381-2323-3/20/\$31.00 ©2020 IEEE}}

\begin{document}

\title{LULC classification by semantic segmentation of satellite images using FastFCN}

\author{
    \IEEEauthorblockN{\textcolor{white}{.}}
     \hfill
    \and
    \IEEEauthorblockN{Md. Saif Hassan Onim$^1$, Aiman Rafeed Bin Ehtesham$^2$, Amreen Anbar$^3$, \\A. K. M. Nazrul Islam$^4$, A. K. M. Mahbubur Rahman$^5$
    \\ \vspace*{-4mm}
    $^{1,2,3,4}$Military Institute of Science \& Technology (MIST), Dhaka - 1216, Bangladesh}\\ 
    $^5$Independent University, Bangladesh (IUB), Dhaka - 1229, Bangladesh\\[0.7mm]
    \small{ \texttt{saif@eece.mist.ac.bd; rafeedaiman@gmail.com; amreen@eece.mist.ac.bd} }
    \hfill
    \and
    \IEEEauthorblockN{\textcolor{white}{.}}
}

\begin{comment}
\\[1cm]
\normalsize{Md. Saif Hassan Onim$^{1}$, Aiman Rafeed Bin Ehesham$^{2}$, Amreen Anbar$^{3}$, A. K. M. Nazrul Islam$^{4}$\\
Department of Electrical, Electronic \& Communication Engineering (EECE),\\
Military Institute of Science \& Technology (MIST), Mirpur Cantonment, Dhaka - 1216, Bangladesh\\
A. K. M. Mahbubur Rahman$^{5}$\\
Department of Computer Science \& Engineering (CSE)\\
Independent University, Aftab Uddin Ahmed Rd, Dhaka - 1229, Bangladesh\\

Email : $^{1}$saif@eece.mist.ac.bd, $^{2}$rafeedaiman@gmail.com, $^{3}$amreen@eece.mist.ac.bd, $^{4}$nazrul@eece.mist.ac.bd, $^{5}$akmmrahman@iub.edu.bd
}
\end{comment}

\maketitle

\begin{abstract}
This paper analyses how well a Fast Fully Convolutional Network (FastFCN) semantically segments satellite images and thus classifies Land Use/Land Cover(LULC) classes. FastFCN was used on Gaofen-2 Image Dataset (GID-2) to segment them in five different classes: BuiltUp, Meadow, Farmland, Water and Forest. The results showed better accuracy (0.93), precision (0.99), recall (0.98) and mean Intersection over Union (mIoU) (0.97) than other approaches like using FCN-8 or eCognition, a readily available software. We presented a comparison between the results. We propose FastFCN to be both faster and more accurate automated method than other existing methods for LULC classification.% The code is available in \texttt{\url{https://github.com/Onimee58/jpu_in_sat_image_segmentation/tree/master/encoding}}

\end{abstract}

\IEEEpeerreviewmaketitle

% RGB images only
% literature shobdo ta thik achhe kina
% k ki korse oita add korte hobe
% GID er info er reference dite hobe
\section{Introduction}
In computer vision (CV), semantic segmentation (SS) \cite{caesar2018coco, mottaghi2014role, zhou2017scene} refers to the assignment of every pixel of an image to a particular class, represented by different colors. Fully convolutional network (FCN) \cite{long2015fully} is a modern approach to that end. Different versions of FCNs have been proposed throughout the literature which have given results of different levels of accuracy, going through different computational complexities.

In our paper, we semantically segmented satellite images of GID-2. The method we have used to segment, namely FastFCN, was proposed by Huikai Wu \emph{et al.} \cite{wu2019fastfcn}. FastFCN outperformed other FCNs in various ways. It consists of the orignal FCN as the backbone and replaces the dilated convolutions \cite{chen2017deeplab} by a novel Joint Pyramid Upsampling (JPU) module. 

The motivation of our research was to find how well FastFCN segments satellite images of the GID-2 dataset into LULC classes. We use the term Land Use and Land Cover (LULC) to refer to the vegetation, water and various other natural or artificial features of the land surface. The five classes considered are: builtup (human built constructions), meadow, farmland, forest and water. Any portion of the image that were not recognized by the FastFCN to be one of these  classes was labeled as ``Unrecognized". 

While it has already been shown to be an efficient method for SS, we still test it with satellite images for two reasons:
(1) SS methods consider the shapes of objects/classes to be segmented. But LULC classes like water/river or forests are not necessarily determined by shapes only. They are determined only by colors and textures.
(2) We wanted to find an automated LULC classification method that works better than the existing ones.

We determined values of different performance parameters (e.g. accuracy \& Intersection over Union (IoU) ) for each classes for every image, compared the overall results with other existing methods and tried to infer from it how good of a replacement can it be for (a) existing software and (b) other algorithms/methods. The results obtained were commendable for most classes with a mean accuracy of 0.93 and mIoU of 0.97.

In short, our contributions are as follows: \textbf{(1)} We proposed a time and memory efficient, automated LULC classification technique that can replace arduous manual method (and the associated costs) without any significant loss of accuracy. \textbf{(2)} We analyzed and made a record of how a well-known segmentation method i.e. FastFCN can perform SS of satellite images. The results of our analysis can work as a future reference of information for studies related to both LULC and SS.

\begin{figure*}
    \centering
    
    \includegraphics[width=\linewidth]{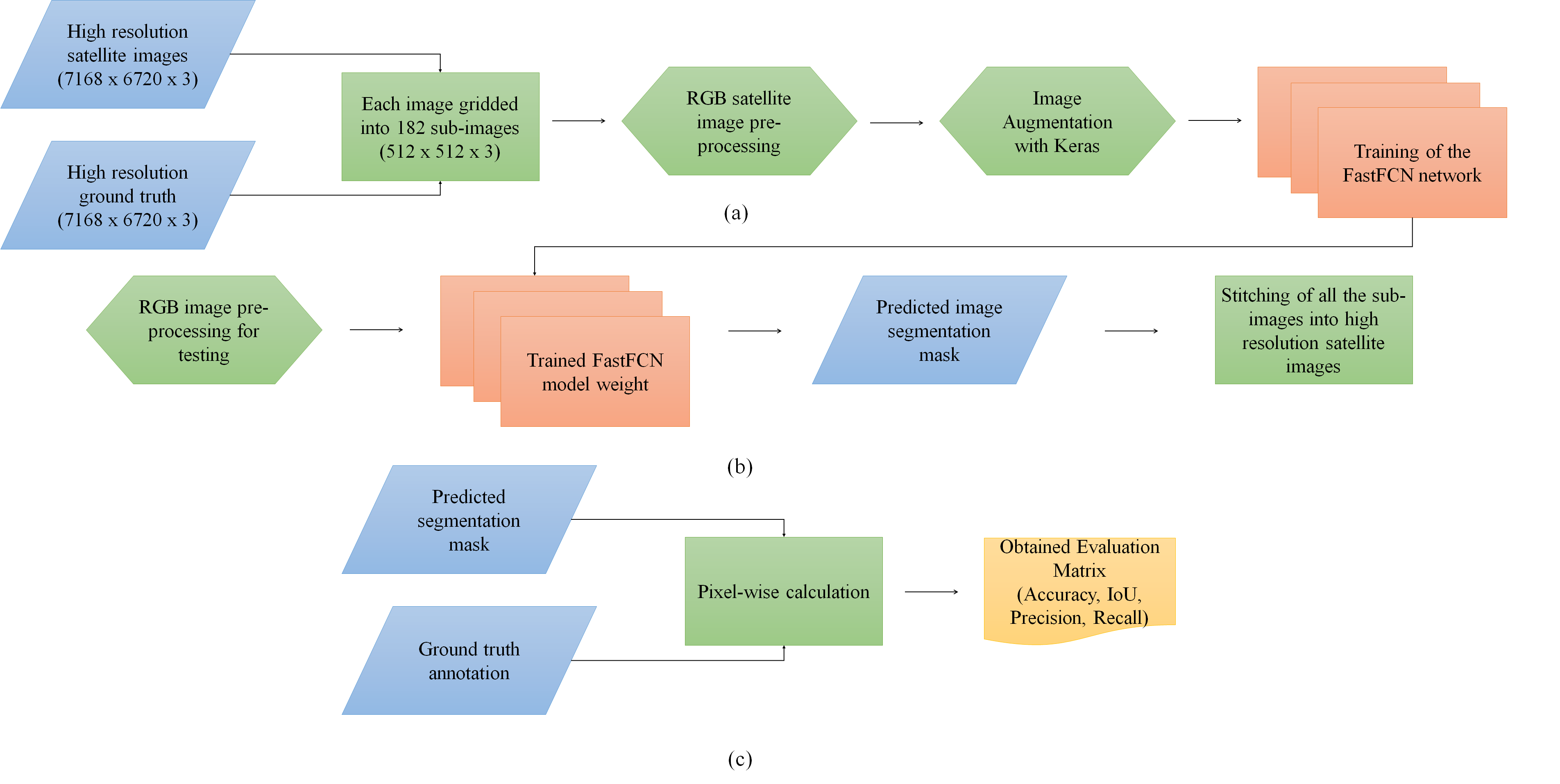}
    
    \caption{Flowchart of developed segmentation process. (a) Training; (b) Testing; (c) Evaluating.}
    \label{fig:flowchart} 
\end{figure*}

\section{Literature Review}

To monitor the limited natural resources available, to detect changes, to plan and introduce new infrastructures, amenities and efficient management of land overall \cite{cheng2015effective, bioucas2012hyperspectral}, LULC classification can be taken as the base.

For performing this classification automatically on satellite images for feature extraction, Cheng \emph{et al.} \cite{cheng2015effective} used Histogram of Oriented Gradients (HOG), Local Binary Pattern (LBP) and Scale-Invariant Feature Transform (SIFT) and as a classifier they used Support Vector Machine (SVM). In processing Geo-special data, deep learning methods have been widely used for land cover classification and segmentation as projected by some of the recent notable works by the researchers with promising results \cite{hamida2017deep, piramanayagam2018supervised}. DenseNet \cite{volpi2016dense} and SegNet\cite{badrinarayanan2017segnet} was used by Ben Hamida \emph{et al.} \cite{hamida2017deep} respectively for fine segmentation and coarse segmentation on multispectral Sentinel-2 images. A fusion of features in a neural architecture (FCN \cite{long2015fully}) was introduced by Piramanayagam \emph{et al.} \cite{piramanayagam2018supervised} for classifying satellite or multisensory aerial images. These performances of automatic LULC classification are not satisfactory due to some limitations.

To overcome these, A. B. S. Nayem \emph{et al.} \cite{2008.10736} introduced  a non-overlapping grid-based approach to train FCN-8 having the architecture of VGG-16. Dividing the full sized satellite images into 224$\times$224 non-overlapping sub-images with a view to preserving the resolution of the input images to be fed to the FCN-8. The model obtained average accuracy of 91.0\% and average IoU of 0.84 which is significant compared to the results obtained from eCognition mentioned before.

%\begin{comment} % tabol
\begin{table*}%[!t]
    \caption{Performance parameter values of the example test result of FastFCN in Fig. \ref{fig:ex1resFast} and of eCognition in Fig. \ref{fig:ex1resEcog}.}
    \centering
    \resizebox{\textwidth}{!}{
    
    \begin{tabular}{|c|c|c|c|c|c|c||c|c|c|c|c|c|}
    \hline 
    \multirow{2}{2em}{Metric} & \multicolumn{6}{|c||}{FastFCN} & \multicolumn{6}{|c|}{eCognition}\\
    \cline{2-13}
     & Weighted & Forest & BuiltUp & Water & Farmland & Meadow & Weighted &	 Forest & BuiltUp & Water & Farmland & Meadow\\
    \hline \hline
    
    Percentage(\%)
    & 100 & 0.44 & 2.83 & 24.59 & 22.64 & 0.00
    & 100 & 0.44 & 2.83 & 24.59 & 22.64 & 0.00\\
    
    Accuracy
    & 1 & 1 & 1 & 0.99 & 1 & 1 
    & 0.70 & 0.80 & 0.98 & 0.90 & 0.83 & 1\\
    
    IoU
    & 0.99 & 0.59 & 0.96 & 0.98 & 0.99 & 1 
    & 0.81 & 0.02 & 0.54 & 0.71 & 0.58 & 1\\
    
    Precision
    & 0.99 & 0.59 & 0.96 & 1 & 0.99 &	1
    & 0.81 & 0.02 & 0.54 & 0.72 & 0.58 & 1\\
    
    Recall
    & 0.99 & 1 & 1 & 0.98 & 1 & 1
    & 0.51 & 1 & 1 & 0.98 & 1 & 1\\
    \hline
    \end{tabular}
    }
     \label{tab:PP for the example}
\end{table*}
% \end{comment}  % tabol

\section{Methodology}
\subsection{Description of Dataset}
The GID dataset is a land cover dataset built using images taken by the Gaofen-2 satellite, an optical satellite among a series of Chinese civilian remote sensing satellites.

The dataset contains 150 tiff images of resolution 7168$\times$6720, providing a spectral range of blue ($0.45-0.52 \mu m$), green ($0.52-0.59 \mu m$), red ($0.63-0.69 \mu m$) and near infrared ($0.77-0.89 \mu m$). The spatial dimension of 7168$\times$6720 covers a geographic area of 506 $km^2$. The dataset can be found at this link : \href{https://whueducn-my.sharepoint.com/:f:/r/personal/xinyi_tong_whu_edu_cn/Documents/GID/Large-scale\%20Classification_5classes?csf=1&web=1&e=zehHQO}{Download}.

\subsubsection{Ground Truths (GT)}
Five different colored pixels represent five different LULC classes in 150 GT images. If a pixel could not be classified as any of the classes under consideration, it has been colored black. These unrecognized data were not considered in our experiments.

\begin{comment}
Fig. \ref{fig:Example} shows an example image and its GT.
\begin{figure}[h]  %subfigg
    \centering
  
    \subfloat[Original image\label{fig:exampleImage}]{%
       \includegraphics[width=0.49\linewidth]{check/1orig.png}}
    \hfill
    \subfloat[Ground truth of the image\label{fig:exampleGT}]{%
        \includegraphics[width=0.49\linewidth]{check/1gt.png}}
    
    \caption{Example image of GID-2.}
    \label{fig:Example} 
\end{figure}
\end{comment}

\subsubsection{Preprocessing}
The original implementation of FastFCN downsamples the input images to fit the input size of ResNet-101 i.e. 512$\times$512$\times$3. This results in loss of information of the original images. So, we cropped each of the 150 original and GT images of size 7168$\times$6720 to 182 sub-images of size 512$\times$512$\times$3 to match the input size of ResNet-101 (the neural network (NN) we used) totalling in 27300 original and GT images. This safely preserves the pixel information of every image.

\noindent
\begin{tabular}{lr}
Training sub-images : 21840 & Testing sub-images : 5460
\end{tabular}

\subsubsection{Data Augmentation}
Each of the images used were converted into 10 different images including the original image as different augmentation methods were used. The images were flipped vertically, horizontally and rotated anti-clockwise 90$^{\circ}$, 180$^{\circ}$ and 270$^{\circ}$.

\subsection{Overall Segmentation Process}
    
    \subsubsection{Training}
        Following preprocessing and data augmentation, we inserted the 512$\times$512$\times$3 training images along with their ground truths into the NN. The training was not done on binary classes rather all 5 LULC classes were trained at once. From the original algorithm we removed the object detection part and the object lists as they were only redundant for satellite images. Fig. \ref{fig:flowchart}\textcolor{red}{a} shows the training procedure in details. The hyperparameters were tuned as - batch size : 16, head above ResNet-101 : JPU, segmentation loss : initialized according to Pytorch, number of epochs : 50, learning rate : 0.01, auxiliary weightloss : 0.2, weight decay : $10^{-4}$.
    \subsubsection{Testing}
        After training, the test images were fed into the NN. The predicted segmentation masks were stitched back together to create high resolution images (Fig. \ref{fig:flowchart}\textcolor{red}{b}).
    \subsubsection{Evaluating}
        All the values of evaluation metrics were calculated from predicted segmentation masks and the ground truths. The calculation was performed pixel-wise to ensure maximum authenticity (Fig. \ref{fig:flowchart}\textcolor{red}{c}).

\subsection{FastFCN}
FCN, the NN transformed from the CNN had limitations from the view point of accuracy of prediction. 
\begin{comment}
The final feature map of reduced spatial resolution led to such performance. With a view to obtaining a final feature map of higher resolution, keeping FCN at the base, evolution led to DilatedFCN followed by EncoderDecoder method. A novel joint upsampling module, JPU \cite{wu2019fastfcn} incorporated with FCN led to FastFCN that outperformed all the previous algorithms of semantic segmentation, producing increased numerical values of the evaluation metrics.
\end{comment}
In DilatedFCN \cite{long2015fully}, the last two downsampling layers have been removed in order to obtain a high-resolution feature map. This results in increased time and memory complexity. FastFCN approximates the final feature map in a different way that reduces these complexities.

\begin{figure}[h]  
    \centering
    
    \includegraphics[width=\linewidth]{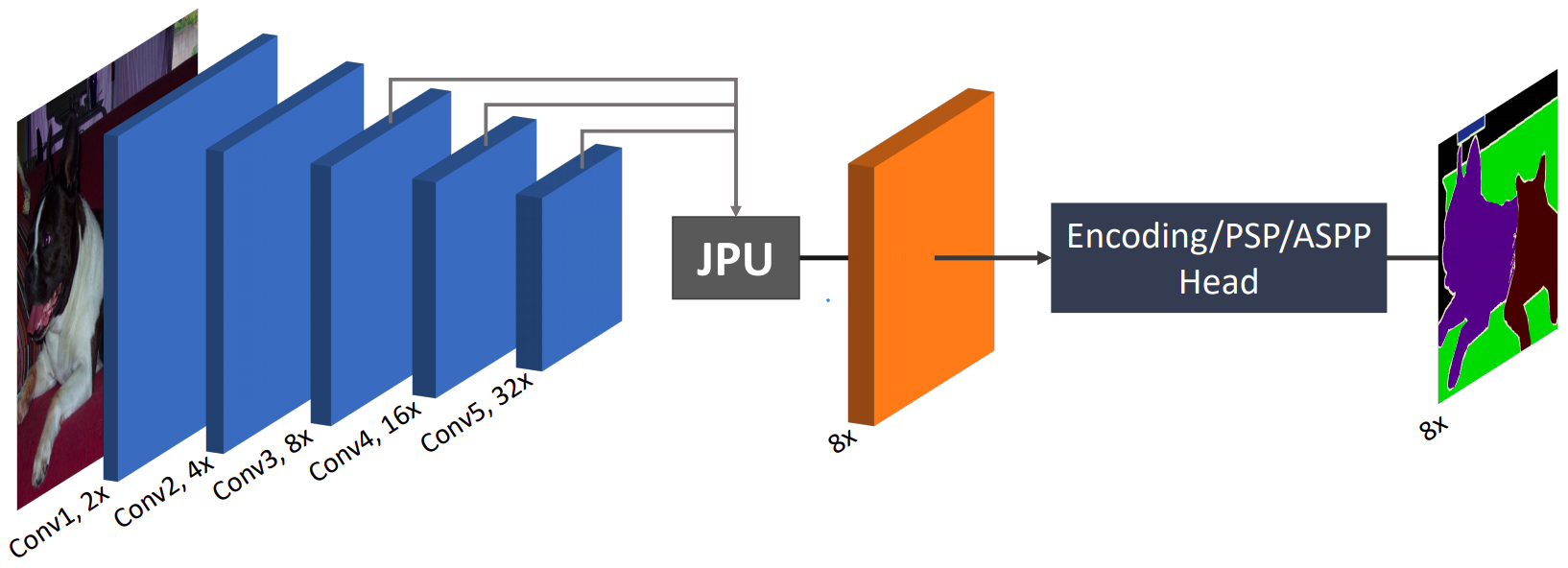}
    
    \caption{Framework of FastFCN \cite{wu2019fastfcn}.}
    \label{fig:fastFCN_framework} 
\end{figure}

In FastFCN, (1) all stride convolution layers removed in DilatedFCN have been restored; (2) each dilated convolution is replaced with a regular convolution. The backbone of FastFCN is the same as the original FCN. The spatial resolution of the five feature maps halves in each layer (Conv1-Conv5), as shown in Fig. \ref{fig:fastFCN_framework}. FastFCN introduces a novel upsampling module that takes the last three feature maps (Conv3-Conv5) as inputs, thus is named Joint Pyramid Upsampling (JPU).

The output feature map $y_s$ is generated as follows:
\begin{eqnarray*}
y_s &=& x \rightarrow C_s \rightarrow C_r \rightarrow \dots \rightarrow C_r\\
&=& x \rightarrow C_r \rightarrow R \rightarrow C_r  \rightarrow \dots \rightarrow C_r\\
&=& y_m \rightarrow R \rightarrow C^n_r\\
&=& y^0_m \rightarrow C^n_r
\end{eqnarray*}

$C_r$, $C_d$, and $C_s$ represent a regular, dilated and stride convolution respectively, and $C^n_r$ is $n$ layers of regular convolutions. The aforementioned equations show that $y_s$ and $y_d$ can be obtained with the same function $C^n_r$ with different inputs: $y^0_m$ and $y_m$, where the former is downsampled from the latter \cite{wu2019fastfcn}. The feature map $y$ that approximates $y_d$ can be obtained by:
\begin{eqnarray*}
y_d &=&  \{y^0_m, y^1_m\}  \rightarrow \hat{h} \rightarrow M\\
\text{where,} ~ \hat{h} &=& \text{argmin}_{h \in \mathcal{H} } \| y_s - h(y^0_m)\| \\
y_m &=& x \rightarrow C_r
\end{eqnarray*}

\begin{comment}
\begin{tabular}{lc}
    Batch size & 16\\
    Head above ResNet-101 & JPU \\
    Segmentation loss initialization & Pytorch \\
    Number of epochs & 50 \\
    Learning rate & 0.01 \\
    Auxiliary weight-loss & 0.2 \\
    Weight decay & $10^{-4}$ \\
    
\end{tabular}
\end{comment}

\section{Experimental Results}

%\begin{comment}  % tabol
\begin{table*}% [!t] 
    \caption{Performance of FastFCN \& FCN-8.}
    \centering
    \resizebox{0.70\textwidth}{!}{
    \begin{tabular}{|l|c|c|c|c||c|c|c|c|}
    \hline 
    \multirow{2}{2em}{Classes} & \multicolumn{4}{|c||}{FastFCN} & \multicolumn{4}{|c|}{FCN-8}\\
    \cline{2-9}
    & Accuracy & IoU & Recall & Precision & Accuracy & IoU & Recall & Precision\\
    \hline \hline
    
    Forest & 0.92 & 0.88 & 0.96 & 0.91  
    & 0.92 & 0.85 & 0.57 & 0.90\\
    
    BuiltUp & 0.99 & 0.93 & 0.98 & 0.95 
    & 0.91 & 0.85 & 0.51 & 0.85\\
    
    Water & 0.98 & 0.92 & 0.96 & 0.95  
    & 0.96 & 0.93 & 0.86 & 0.91\\
    
    Farmland & 0.98 & 0.95 & 0.97 & 0.98  
    & 0.85 & 0.74 & 0.71 & 0.70\\
    
    Meadow & 0.93 & 1 & 1 & 1  
    & - & - & - & - \\
    
    Weighted average & 0.93 & 0.97 & 0.98 & 0.99   
    & 0.91 & 0.84 & 0.66 & 0.84\\
    
    \hline
    \end{tabular}
    }
     \label{tab:Overall performance}
\end{table*} % tabol

In this section, we first show the performance of FastFCN in segmenting satellite images and compare with eCognition,

\begin{figure}[h]  %subfigg
    \centering
  
    \subfloat[Original test image\label{fig:ex1orig}]{%
       \includegraphics[width=0.49\linewidth]{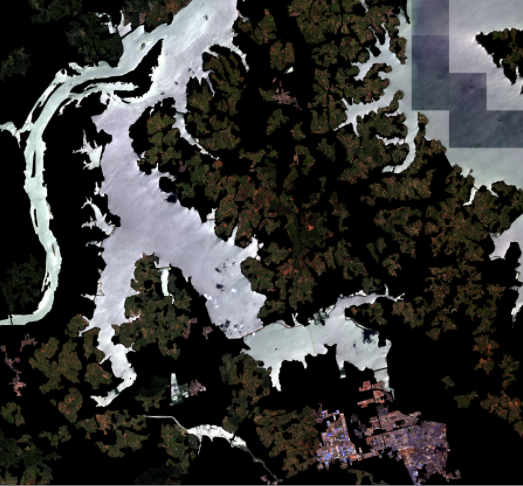}}
    \hfill
    \subfloat[Ground truth\label{fig:ex1gt}]{%
        \includegraphics[width=0.49\linewidth]{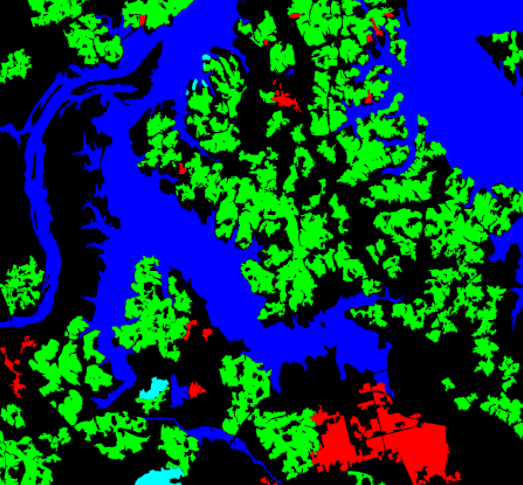}}
    \hfill
    \subfloat[Result (FastFCN)\label{fig:ex1resFast}]{%
        \includegraphics[width=0.49\linewidth]{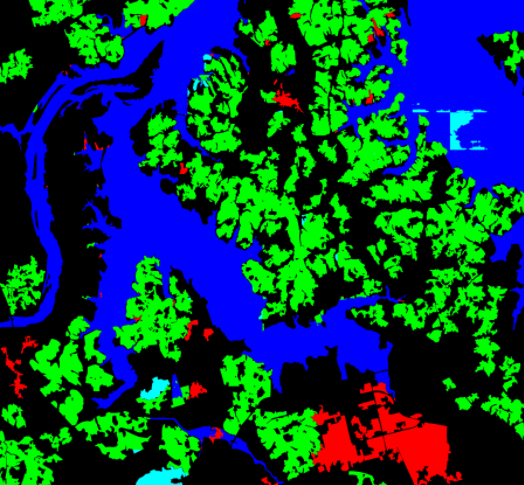}}
    \hfill
    \subfloat[Result (eCognition)\label{fig:ex1resEcog}]{%
        \includegraphics[width=0.49\linewidth]{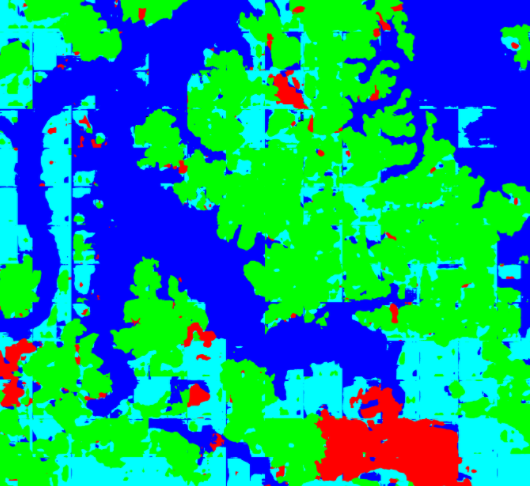}}
    
    \caption{Example test image, its ground truth and test results.}
    \label{fig:Example results} 
\end{figure}

\noindent a Geographic Information System (GIS) software and FCN-8. We then try to analyse why the test results of some classes were more error-prone than others and how we could have obtained more error-free results.

There were 30 test images among the 150 images of the dataset. We have tested the resultant stitched images of this test images based on the values of the following performance parameters : 
(1) Accuracy 
(2) Precision
(3) IoU and
(4) Recall.

IoU (Fig. \ref{fig:IoU}) is one of the most important parameters in image processing. The Intersection over Union (IoU) metric is a method to quantify the percent overlap between the Ground Truth and the predicted output. The IoU can be calculated using the equation:

\begin{figure}[h]  
    \centering
    
    \includegraphics[width=\linewidth]{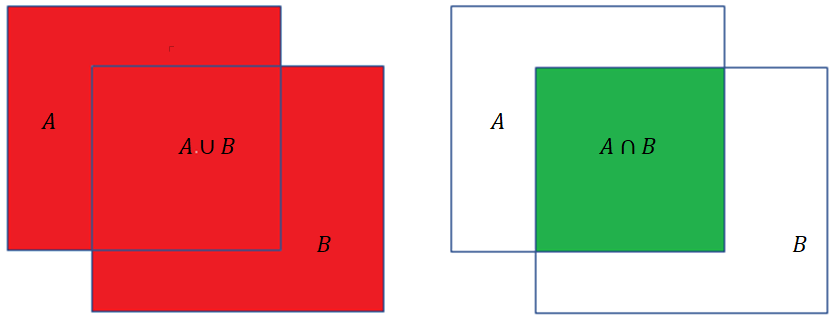}
    
    \caption{Intersection over Union. A \& B represent GT and predicted output respectively.}
    \label{fig:IoU} 
\end{figure}

$$
\text{IoU} = \frac{\text{Intersection}}{\text{Union}} = \frac{\text{TP}}{\text{TP}+\text{FP}+\text{FN}}\\
$$

Where, TP $=$ True Positive, FP = False Positive and FN = False Negative.

\begin{figure}[h]  %subfigg
    \centering
 
    \subfloat[Ground Truth\label{fig:collage1}]{%
       \includegraphics[width=0.49\linewidth]{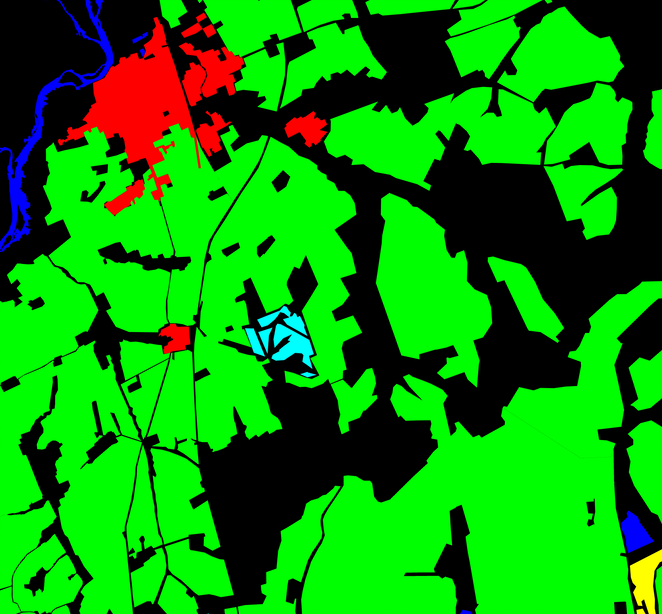}}
    \hfill
    \subfloat[Test result\label{fig:collage2}]{%
        \includegraphics[width=0.49\linewidth]{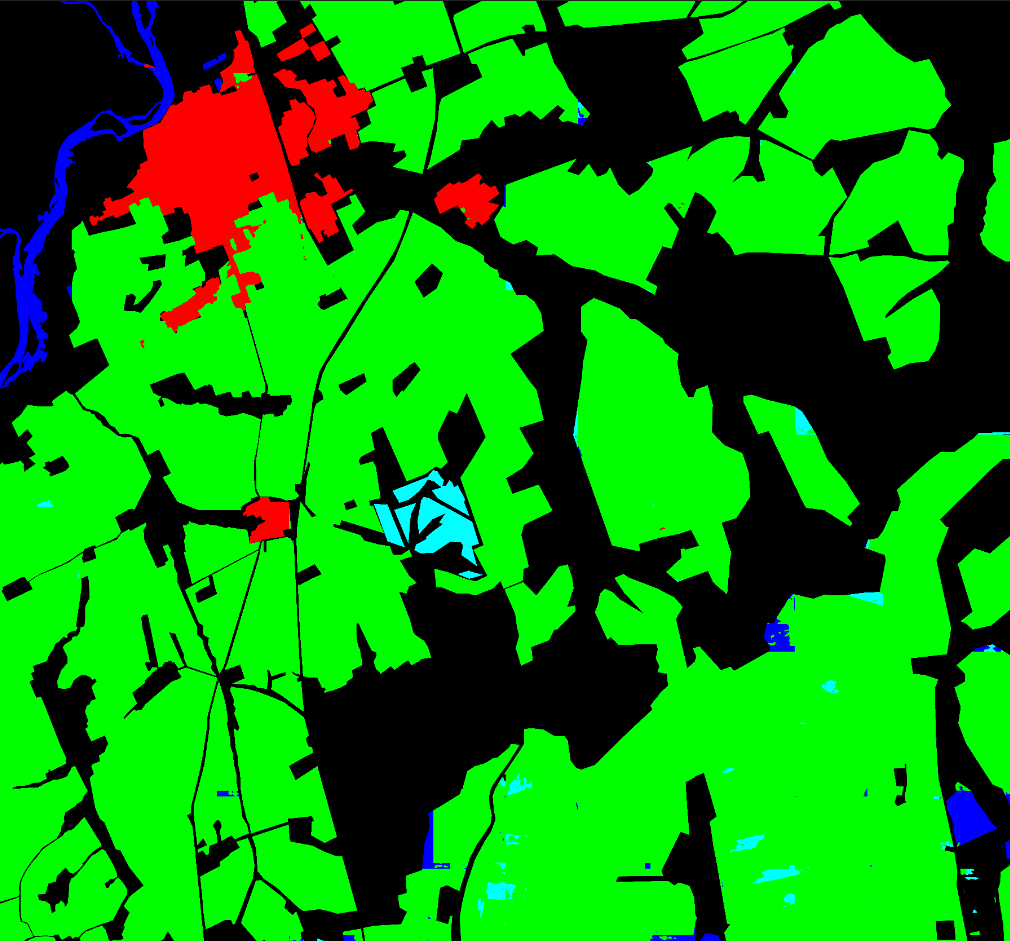}}
    \hfill
    \subfloat[Error\label{fig:collage3}]{%
        \includegraphics[width=0.98\linewidth]{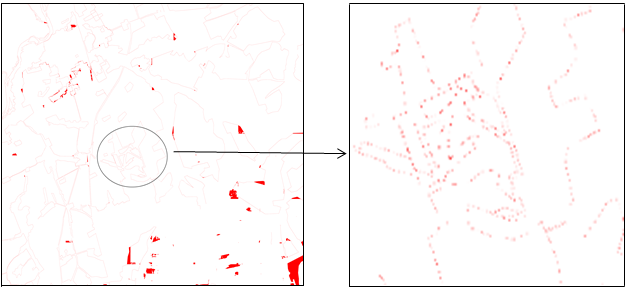}}
    
    \caption{An example image where the classes have sharp edges}
    \label{fig:Collages} 
\end{figure}

\begin{figure}[h]  %subfigg
    \centering
  
    \subfloat[Original image\label{fig:forestMeadow A}]{%
       \includegraphics[width=0.49\linewidth]{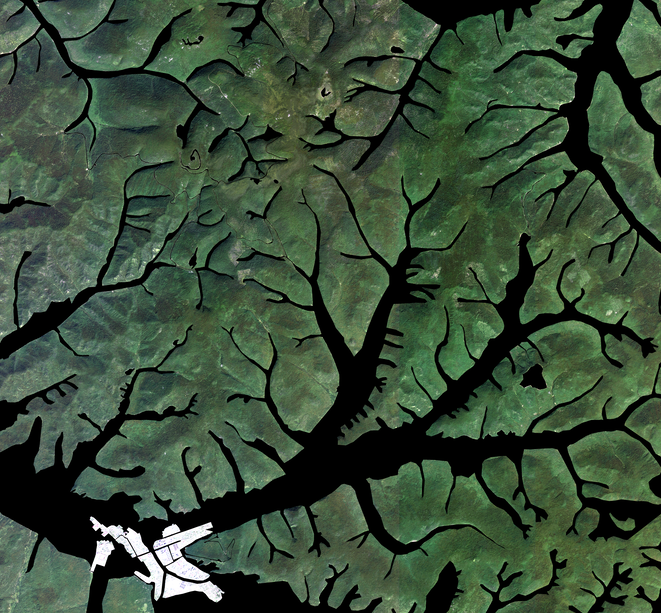}}
    \hfill
    \subfloat[Ground Truth\label{fig:forestMeadow B}]{%
        \includegraphics[width=0.49\linewidth]{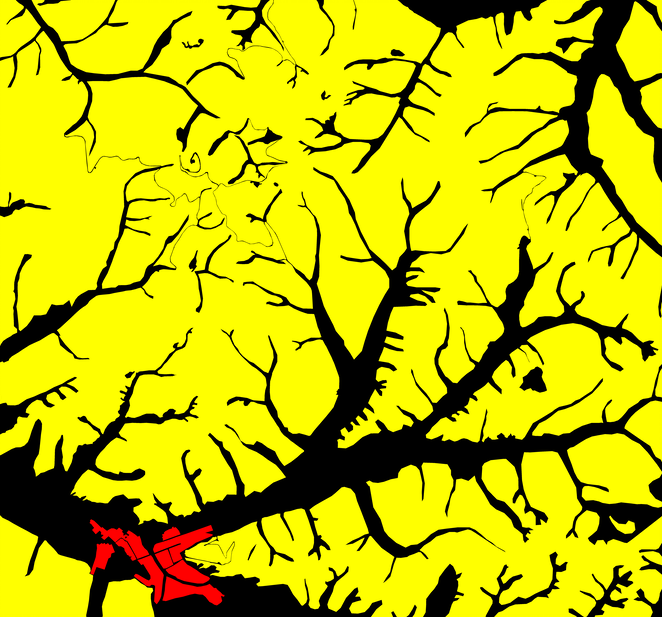}}
    \hfill
    \subfloat[Test result\label{fig:forestMeadow C}]{%
        \includegraphics[width=0.49\linewidth]{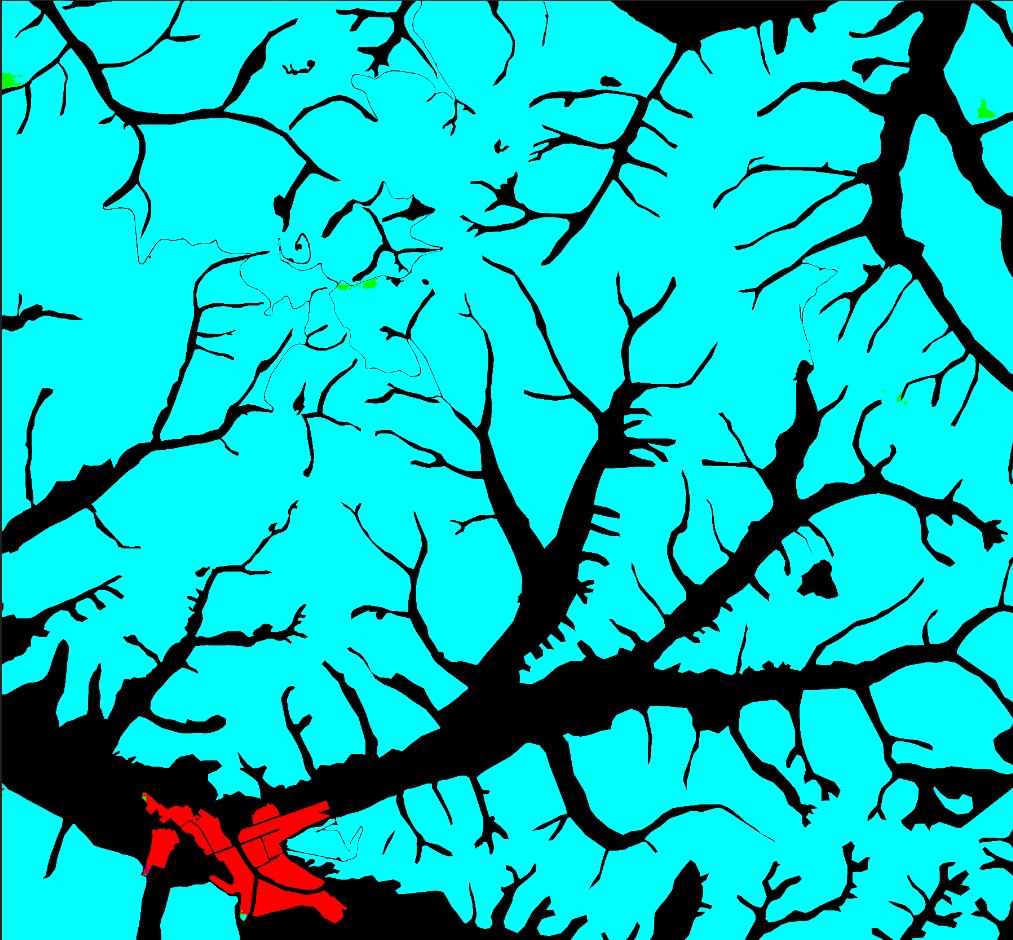}}
    \hfill
    \subfloat[Error\label{fig:forestMeadow D}]{%
        \includegraphics[width=0.49\linewidth]{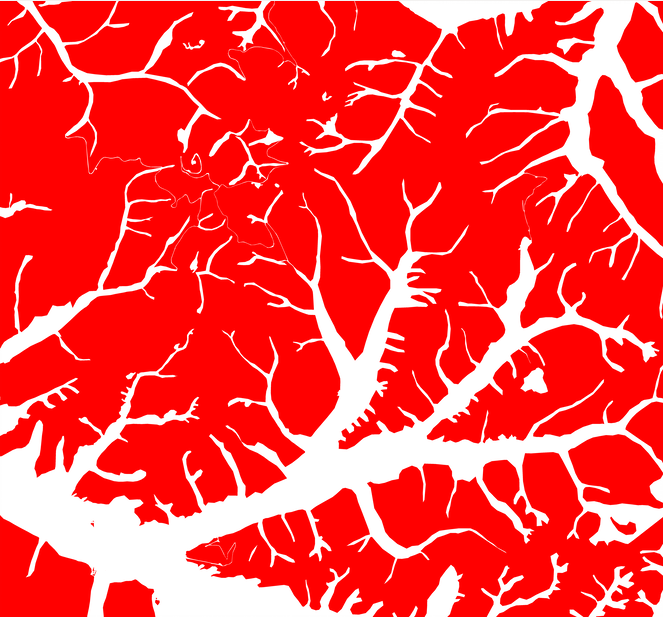}}
    
    \caption{(a) The original image; (b) A large portion of the image is 
    ``meadow" according to the GT; (c) The test result, the entirety of the meadow class has been classified to be forest; (d) The erroneous portion of the test result, shown in red.}
    \label{fig:forestMeadow} 
\end{figure}

The performance parameters of each class of each image have been calculated separately, then averaged over 30 images to get the overall performance of FastFCN.

Different classes have been represented by different colors:\\[2pt]
\begin{tabular}{llll}
\textbf{(a)} Unrecognized&: Black & \textbf{(b)} Farmland &: Green \\
\textbf{(c)} Water &: Blue & \textbf{(d)}  Meadow   &: Yellow \\
\textbf{(e)} BuiltUp  &: Red & \textbf{(f)} \hspace{2pt}Forest   &: Cyan \\
\end{tabular}

\subsection{Test Results and Comparison}

An example test image has been randomly picked to compare its results with eCognition. Its GT and test result from FastFCN and eCognition have been shown in Fig.\ref{fig:Example results}. The performance parameter values of the test results are given in Table \ref{tab:PP for the example}.

Table \ref{tab:Overall performance} shows the overall test result of FastFCN averaged over the 30 test images. We show results of a similar experiment of semantic segmentation of satellite images by A. B. S. Nayem \emph{et al.} \cite{2008.10736} alongside ours. FastFCN outperformed FCN-8 in classifying all the classes.

\subsection{Error Analysis}
In this section, we will explore where ResNet-101 failed to segment and why. In Fig. \ref{fig:collage3} and Fig. \ref{fig:forestMeadow D}, the true positives are represented by white and errors in classification are represented by red.

In maximum cases, the NN faced difficulties with sharp edges. Where the region is smaller than $10m^2$ ground area, the FastFCN could no longer detect the class. Zooming into the high resolution images, we can identify the sharp edges having false positive segmentation.

%In Fig. \ref{fig:collage3}, an example is shown. 
Fig. \ref{fig:collage1} and \ref{fig:collage2}, respectively, are the GT and test result of an image in which various classes have sharp edges. The error in classification is shown in Fig. \ref{fig:collage3}, where a zoomed in portion of the image is shown for clarity. We see that the misclassified portion is predominantly the sharp edges. 
% The sharp edges of the classes are clearly visible in the result/GT image \textcolor{blue}{[]} and the errors occurred mostly on those regions \textcolor{blue}{[]}.

Another reason for errors in segmentation was due to unbalanced training of the classes. Training was biased to the classes which were present considerably more in the training images. If all the classes were trained equally, the performance might have been better.

We see that, the results for the forest and the meadow classes are significantly worse than the other classes. This is probably due to the similarities of textures and colors between the two classes for which the NN confused one to be another in the testing process. An example of this is shown in the Fig. \ref{fig:forestMeadow}.

\section{Conclusion}

In this paper, we have attempted to analyze how a modern image segmentation method (i.e.  FastFCN) performs in semantic segmentation of satellite images. Considering the results, we proposed it to be a great automated method for LULC classification. We compared its performance with another method (i.e. FCN-8) and found it to be better than that of FCN-8’s in several regards. The results showed that some of the LULC classes, namely ``meadow" and ``forest", were not segmented as accurately as others. Also, the classes were not accurately detected at their boundaries. We hope to resolve these issues and work with bigger datasets in our future work.

%\begin{comment}
\section*{Acknowledgment}
This project is supported by a grant from the Bangladesh Information and Communication Technology
Ministries ICT Division, and Independent University, Bangladesh (IUB). We also thank the Department of Electrical, Electronic \& Communication Engineering (EECE) of MIST for their support.
%\end{comment}

% Generated by IEEEtran.bst, version: 1.12 (2007/01/11)

\end{document}